%% file: main.tex
\documentclass[10pt,twocolumn,letterpaper]{article}

\input{package}
\input{macro}

\iccvfinalcopy %

\def\iccvPaperID{751} %

\ificcvfinal\fi
\begin{document}

\title{TSM: Temporal Shift Module for Efficient Video Understanding} 

\author{Ji Lin\\
MIT\\
{\tt\small jilin@mit.edu}
\and
Chuang Gan\\
MIT-IBM Watson AI Lab\\
{\tt\small ganchuang@csail.mit.edu}
\and
Song Han\\
MIT\\
{\tt\small songhan@mit.edu}
}

\maketitle

\input{text/0_abstract}

\input{text/1_introduction}

\input{text/2_related_work}

\input{text/3_tsm_module}

\input{text/4_tsm_video_network}
\input{text/5_experiments}

\input{text/6_conclusion}
\ificcvfinal\input{text/7_acknowledge}\fi

{\small
\bibliographystyle{ieee_fullname}
\bibliography{egbib.bib}
}

\input{text/appendix}

\end{document}

%% file: package.tex
\usepackage{color,xcolor}
\usepackage{epsfig}
\usepackage{graphicx}
\usepackage{subfiles}

\usepackage{adjustbox}
\usepackage{array}
\usepackage{booktabs}
\usepackage{colortbl}
\usepackage{float,wrapfig}
\usepackage{hhline}
\usepackage{multirow}
\usepackage{subcaption} %
\usepackage[labelfont={bf},labelsep={period},font={small}]{caption}

\let\llncssubparagraph\subparagraph
\let\subparagraph\paragraph
\usepackage[compact]{titlesec}
\let\subparagraph\llncssubparagraph

\usepackage{amsmath,amsfonts,amssymb}
\usepackage{bm}
\usepackage{nicefrac}
\usepackage{microtype}

\usepackage{changepage}
\usepackage{extramarks}
\usepackage{fancyhdr}
\usepackage{lastpage}
\usepackage{setspace}
\usepackage{soul}
\usepackage{xspace}

\usepackage{url}

\usepackage{algorithm}
\usepackage{algpseudocode}
\usepackage{enumerate}

\usepackage{iccv}
\usepackage{times}

\usepackage[pagebackref=true,breaklinks=true,letterpaper=true,colorlinks,bookmarks=false]{hyperref}

\usepackage{pbox}
\usepackage{footnote}
\usepackage{tablefootnote}

\usepackage{paralist}

\usepackage{enumitem}
\setitemize{noitemsep,topsep=0pt,parsep=0pt,partopsep=0pt}

%% file: macro.tex
\usepackage{paralist}
\newcommand{\netFull}{TSM\xspace}
\newcommand{\netHead}{TSM}

\newcommand{\method}{Temporal Shift\xspace}

\newcommand{\myparagraph}[1]{\vspace{-3pt}\paragraph{#1}}

\makeatletter
\newcommand\footnoteref[1]{\protected@xdef\@thefnmark{\ref{#1}}\@footnotemark}
\makeatother

%% file: text/0_abstract.tex
\begin{abstract}
 
The explosive growth in video streaming gives rise to challenges on performing video understanding at high accuracy and low computation cost. 
Conventional 2D CNNs are computationally cheap but cannot capture temporal relationships; 3D CNN based methods can achieve good performance but are computationally intensive, making it expensive to deploy. 
In this paper, we propose a generic and effective Temporal Shift Module (TSM) that enjoys both high efficiency and high performance. Specifically, it can achieve the performance of 3D CNN but maintain 2D CNN's complexity. TSM shifts part of the channels along the temporal dimension; thus facilitate information exchanged among neighboring frames. It can be inserted into 2D CNNs to achieve temporal modeling at zero computation and zero parameters.  We also extended TSM to online setting, which enables real-time low-latency online video recognition and video object detection. TSM is accurate and efficient: it ranks the first place on the Something-Something leaderboard upon publication; on Jetson Nano and Galaxy Note8, it achieves a low latency of 13ms and 35ms for online video recognition. 
The code is available at: \url{https://github.com/mit-han-lab/temporal-shift-module}.
\end{abstract}

%% file: text/1_introduction.tex
\section{Introduction}

Hardware-efficient video understanding is an important step towards real-world deployment, both on the cloud and on the edge. For example, there are over $10^5$ hours of videos uploaded to YouTube every day to be processed for recommendation and ads ranking; tera-bytes of sensitive videos in hospitals need to be processed locally on edge devices to protect privacy. 
All these industry applications require both accurate and efficient video understanding.

\input{figureText/fig_tsm_module_new.tex}

Deep learning has become the standard for video understanding over the years~\cite{tran2015learning, wang2016temporal, carreira2017quo, wang2017non, zolfaghari2018eco, xie2018rethinking, zhou2017temporal}. One key difference between video recognition and image recognition is the need for \emph{temporal modeling}. For example, to distinguish between opening and closing a box, reversing the order will give opposite results, so temporal modeling is critical. Existing efficient video understanding approaches directly use 2D CNN~\cite{karpathy2014large, simonyan2014two, wang2016temporal, zhou2017temporal}. However, 2D CNN on individual frames cannot well model the temporal information.
3D CNNs~\cite{tran2015learning, carreira2017quo} can jointly learn spatial and temporal features but the computation cost is large, making the deployment on edge devices difficult; it cannot be applied to real-time online video recognition. 
There are works to trade off between temporal modeling and computation, such as post-hoc fusion~\cite{girdhar2017actionvlad, feichtenhofer2016convolutional, zhou2017temporal, donahue2015long} and mid-level temporal fusion~\cite{zolfaghari2018eco, xie2018rethinking, tran2018closer}. Such methods sacrifice the low-level temporal modeling for efficiency, but much of the useful information is lost during the feature extraction before the temporal fusion happens.

In this paper, we propose a new perspective for efficient temporal modeling in video understanding by proposing a novel Temporal Shift Module (TSM). Concretely, an activation in a video model can be represented as $A\in \mathbb{R}^{N\times C\times T \times H \times W}$,  where $N$ is the batch size, $C$ is the number of channels, $T$ is the temporal dimension, $H$ and $W$ are the spatial resolutions. Traditional 2D CNNs operate independently over the dimension $T$; thus no temporal modeling takes effects (Figure~\ref{fig:shift_ori}).  In contrast, our Temporal Shift Module (TSM) shifts the channels along the temporal dimension, both forward and backward. As shown in Figure~\ref{fig:shift_zero}, the information from neighboring frames is mingled with the current frame after shifting. Our intuition is: the convolution operation consists of \emph{shift} and \emph{multiply-accumulate}. We \emph{shift} in the time dimension by $\pm1$ and fold the \emph{multiply-accumulate} from time dimension to channel dimension. For real-time online video understanding, future frames can't get shifted to the present, so we use a uni-directional TSM (Figure~\ref{fig:shift_online}) to perform online video understanding. 

 \vspace{5pt}
Despite the zero-computation nature of the shift operation, we empirically find that simply adopting the spatial shift strategy~\cite{wu2017shift} used in image classifications introduces two major issues for video understanding:
(1) it is \emph{not efficient}: shift operation is conceptually zero FLOP but incurs data movement. The additional cost of data movement is non-negligible and will result in latency increase. This phenomenon has been exacerbated in the video networks since they usually have a large memory consumption (5D activation). 
(2) It is \emph{not accurate}: shifting too many channels in a network will significantly hurt the spatial modeling ability and result in performance degradation.  To tackle the problems, we make two technical contributions. (1) We use a \emph{temporal partial shift} strategy: instead of shifting all the channels, we shift only a small portion of the channels for efficient temporal fusion. Such strategy significantly cuts down the data movement cost (Figure~\ref{fig:tsm_discuss_a}).
 (2) We insert TSM inside \emph{residual branch} rather than outside so that the activation of the current frame is preserved, which does not harm the spatial feature learning capability of the 2D CNN backbone.  

 \vspace{5pt}
 The contributions of our paper are summarized as follows:

\begin{compactitem}
\setlength\itemsep{1em}
\item We provide a new perspective for efficient video model design by temporal shift, which is computationally free but has strong spatio-temporal modeling ability. 

\item We observed that naive shift cannot achieve high efficiency or high performance.
We then proposed two technical modifications \emph{partial shift} and \emph{residual shift} to realize a high efficiency model design.

\item We propose  \emph{bi-directional TSM} for \emph{offline} video understanding that achieves state-of-the-art performance. It ranks the first on Something-Something leaderboard upon publication.

\item We propose \emph{uni-directional TSM} for \emph{online} real-time video recognition with strong temporal modeling capacity at low latency on edge devices.

\end{compactitem}

%% file: figureText/fig_tsm_module_new.tex
\begin{figure}[t]
\centering
\begin{subfigure}[b]{0.149\textwidth}
	\includegraphics[width=\textwidth]{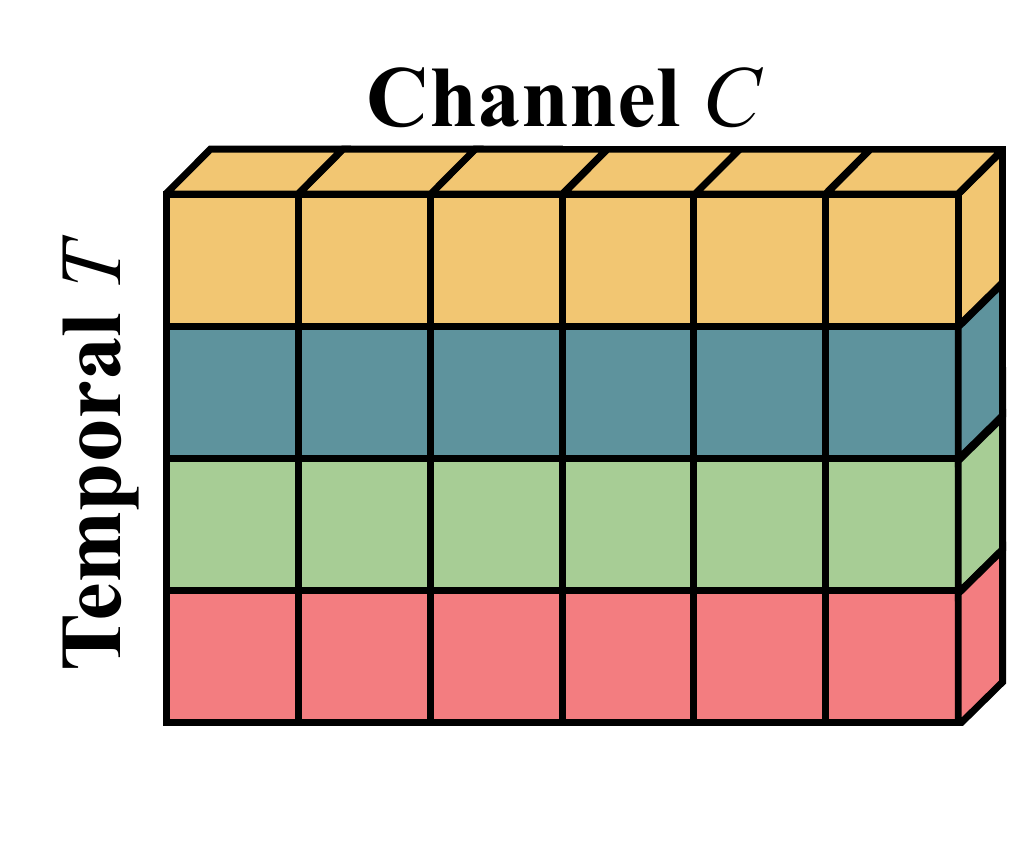}
	\caption{The original tensor without shift.
}
	\label{fig:shift_ori}
\end{subfigure}
~
\begin{subfigure}[b]{0.149\textwidth}
	\includegraphics[width=\textwidth]{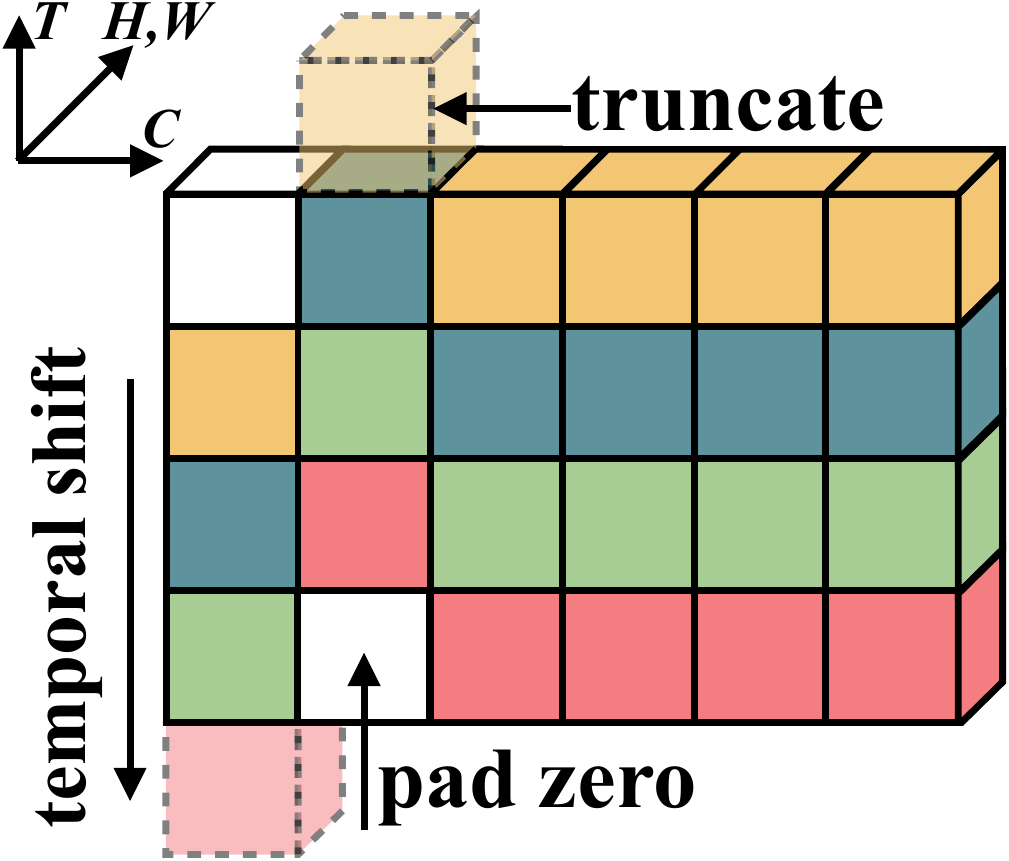}
	\caption{Offline temporal shift (bi-direction). 
}
	\label{fig:shift_zero}
\end{subfigure}
~
\begin{subfigure}[b]{0.149\textwidth}
	\includegraphics[width=\textwidth]{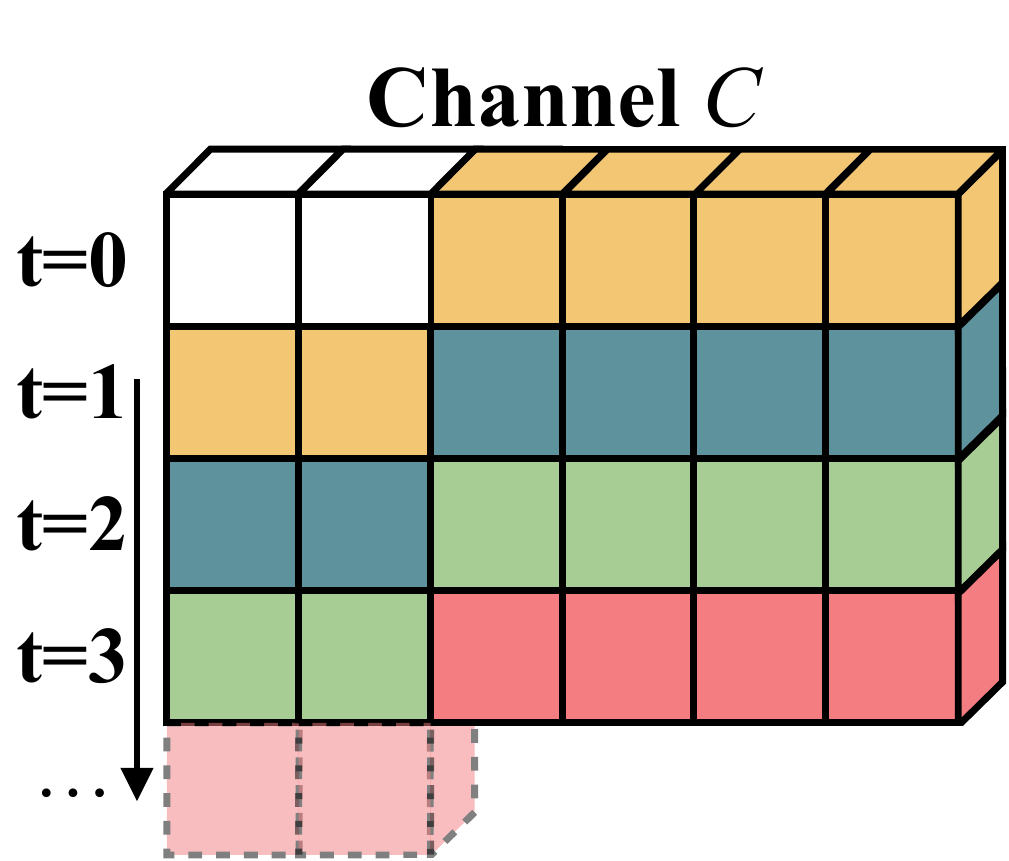}
	\caption{Online temporal shift (uni-direction). 
}
	\label{fig:shift_online}
\end{subfigure}
\caption{\textbf{Temporal Shift Module (TSM)} performs efficient temporal modeling by moving the feature map along the temporal dimension. It is computationally free on top of a 2D convolution, but achieves strong temporal modeling ability.
TSM efficiently supports both \textbf{offline} and \textbf{online} video recognition.
Bi-directional TSM mingles both past and future frames with the current frame, which is suitable for high-throughput offline video recognition. Uni-directional TSM mingles only the past frame with the current frame, which is suitable for low-latency online video recognition. 
}
\vspace{-15pt}
\label{fig:shift}
\end{figure}

%% file: text/2_related_work.tex
\section{Related Work}

\subsection{Deep Video Recognition}

\myparagraph{2D CNN.}
Using the 2D CNN is a straightforward way to conduct video recognition~\cite{karpathy2014large, simonyan2014two, wang2016temporal, gan2015devnet, feichtenhofer2016spatiotemporal, feichtenhofer2016convolutional, bilen2016dynamic}. For example, Simonyan~\etal~\cite{simonyan2014two} designed a two-stream CNN for RGB input (spatial stream) and optical flow~\cite{zach2007duality} input (temporal stream) respectively. Temporal Segment Networks (TSN)~\cite{wang2016temporal} extracted averaged features from strided sampled frames. Such methods are more efficient compared to 3D counterparts but cannot infer the temporal order or more complicated temporal relationships. 

\myparagraph{3D CNN.}
3D convolutional neural networks can jointly learn spatio-temporal features. Tran~\etal~\cite{tran2015learning} proposed a 3D CNN based on VGG models, named C3D, to learn spatio-temporal features from a frame sequence. Carreira and Zisserman~\cite{carreira2017quo} proposed to inflate all the 2D convolution filters in an Inception V1 model~\cite{szegedy2015going} into 3D convolutions. However, 3D CNNs are computationally heavy, making the deployment difficult. They also have more parameters than 2D counterparts, thus are more prone to over-fitting. On the other hand, our \netFull has the same spatial-temporal modeling ability as 3D CNN while enjoying the same computation and parameters as the 2D CNNs.

\myparagraph{Trade-offs.}
There have been attempts to trade off expressiveness and computation costs. Lee~\etal~\cite{lee2018motion} proposed a motion filter to generate spatio-temporal features from 2D CNN.
Tran~\etal~\cite{tran2018closer} and Xie~\etal~\cite{xie2018rethinking} proposed to study mixed 2D and 3D networks, either first using 3D and later 2D (bottom-heavy) or first 2D and later 3D (top-heavy) architecture. ECO~\cite{zolfaghari2018eco} also uses a similar top-heavy architecture to achieve a very efficient framework. Another way to save computation is to decompose the 3D convolution into a 2D spatial convolution and a 1D temporal convolution~\cite{tran2018closer, qiu2017learning, sun2015human}. For mixed 2D-3D CNNs, they still need to remove low-level temporal modeling or high-level temporal modeling. Compared to decomposed convolutions, our method completely removes the computation cost of temporal modeling has enjoys better hardware efficiency.

\subsection{Temporal Modeling}
A direct way for temporal modeling is to use 3D CNN based methods as discussed above. Wang~\etal~\cite{wang2017non} proposed a spatial-temporal non-local module to capture long-range dependencies. Wang~\etal~\cite{wang2018videos} proposed to represent videos as space-time region graphs. 
An alternative way to model the temporal relationships is to use 2D CNN + post-hoc fusion~\cite{girdhar2017actionvlad, feichtenhofer2016convolutional, zhou2017temporal, donahue2015long}. Some works use LSTM~\cite{hochreiter1997long} to aggregate the 2D CNN features~\cite{yue2015beyond, donahue2015long, srivastava2015unsupervised, gan2016webly, gan2016you}.
Attention mechanism also proves to be effective for temporal modeling~\cite{sharma2015action, li2018videolstm, long2018attention}.
Zhou~\etal~\cite{zhou2017temporal} proposed Temporal Relation Network to learn and reason about temporal dependencies.
The former category is computational heavy, while the latter cannot capture the useful low-level information that is lost during feature extraction. Our method offers an efficient solution at the cost of 2D CNNs, while enabling both low-level and high-level temporal modeling, just like 3D-CNN based methods.

\subsection{Efficient Neural Networks}
The efficiency of 2D CNN has been extensively studied. Some works focused on designing an efficient model~\cite{iandola2016squeezenet, howard2017mobilenets, sandler2018mobilenetv2, zhang2017shufflenet}. Recently neural architecture search~\cite{zoph2016neural, zoph2017learning, liu2017progressive} has been introduced to find an efficient architecture automatically~\cite{tan2018mnasnet, cai2018proxylessnas}. Another way is to prune, quantize and compress an existing model for efficient deployment~\cite{han2015learning, han2015deep, lin2017runtime, zhu2016trained, he2018amc, wang2018haq}. Address shift, which is a hardware-friendly primitive, has also been exploited for compact 2D CNN design on image recognition tasks~\cite{wu2017shift, zhong2018shift}. Nevertheless, 
we observe that directly adopting the shift operation on video recognition task neither maintains efficiency nor accuracy, due to the complexity of the video data.

%% file: text/3_tsm_module.tex
\section{\method Module (TSM)}\label{tsm_module}
We first explain the intuition behind TSM: data movement and computation can be separated in a convolution. However, we observe that such naive shift operation neither achieves high efficiency nor high performance. To tackle the problem, we propose two techniques minimizing the data movement and increasing the model capacity, which leads to the efficient TSM module.

\subsection{Intuition}

Let us first consider a normal convolution operation. For brevity, we used a 1-D convolution with the kernel size of 3 as an example. Suppose the weight of the convolution is $W=(w_1, w_2, w_3)$, and the input $X$ is a 1-D vector with infinite length. The convolution operator $Y=\text{Conv}(W, X)$ can be written as:
$ Y_i =  w_1 X_{i-1} + w_2 X_{i} + w_3 X_{i+1}$.
We can decouple the operation of convolution into two steps: \emph{shift} and \emph{multiply-accumulate}: we shift the input $X$ by $-1, 0, +1$ and multiply by $w_1, w_2, w_3$ respectively, which sum up to be $Y$. Formally, the \emph{shift} operation is:
\begin{equation}
\vspace{-5pt}
    X^{-1}_i = X_{i-1},~~~~ X^{0}_i = X_i, ~~~~ X^{+1}_i = x_{i+1}
\end{equation}
and the \emph{multiply-accumulate} operation is:
\begin{equation}
\vspace{-5pt}
    Y=w_1 X^{-1} + w_2 X^{0} + w_3 X^{+1}
\end{equation}
The first step \emph{shift} can be conducted without any multiplication. While the second step is more computationally expensive, our \method module \emph{merges} the \emph{multiply-accumulate} into the following 2D convolution, so 
it introduces no extra cost compared to 2D CNN based models.

The proposed \method module is described in Figure~\ref{fig:shift}.
In Figure~\ref{fig:shift_ori}, we describe a tensor with $C$ channels and $T$ frames. The features at different time stamps are denoted as different colors in each row. 
Along the temporal dimension, we shift part of the channels by $-1$, another part by $+1$, leaving the rest un-shifted (Figure~\ref{fig:shift_zero}).
For online video recognition setting, we also provide an online version of TSM (Figure~\ref{fig:shift_online}). In the online setting, we cannot access future frames, therefore, we only shift from past frames to future frames in a uni-directional fashion.

\subsection{Naive Shift Does Not Work}\label{sec:place_proportion}

Despite the simple philosophy behind the proposed module, we find that directly applying the spatial shift strategy~\cite{wu2017shift} to the temporal dimension cannot provide high performance nor efficiency. To be specific, if we shift all or most of the channels, it brings two disasters:  \textbf{(1) Worse efficiency due to large data movement}. The shift operation enjoys no computation, but it involves data movement. 
Data movement increases the memory footprint and inference latency on hardware. Worse still, such effect is exacerbated in the video understanding networks due to large activation size (5D tensor). When using the naive shift strategy shifting every map, we observe a 13.7\% increase in CPU latency and 12.4\% increase in GPU latency, making the overall inference slow. \textbf{(2) Performance degradation due to worse spatial modeling ability.} By shifting part of the channels to neighboring frames, the information contained in the channels is no longer accessible for the current frame, which may harm the spatial modeling ability of the 2D CNN backbone. We observe a 2.6\% accuracy drop when using the naive shift implementation compared to the 2D CNN baseline (TSN).

\subsection{ Module Design}

\input{figureText/fig_tsm_discuss.tex}

To tackle the two problem from naive shift implementation, we discuss two technical contributions.

\myparagraph{Reducing Data Movement.}
To study the effect of data movement, we first measured the inference latency of TSM models and 2D baseline on different hardware devices. We shifted different proportion of the channels and measured the latency. We measured models with ResNet-50 backbone and 8-frame input using no shift (2D baseline), partial shift ($1/8, 1/4, 1/2$) and all shift (shift all the channels). The timing was measure on server GPU (NVIDIA Tesla P100), mobile GPU (NVIDIA Jetson TX2) and CPU (Intel Xeon E5-2690). We report the average latency from 1000 runs after 200 warm-up runs. We show the overhead of the shift operation as the percentage of the original 2D CNN inference time in~\ref{fig:tsm_discuss_a}. 
We observe the same overhead trend for different devices. If we shift all the channels, the latency overhead takes up to \textbf{13.7\%} of the inference time on CPU, which is definitely \textbf{non-negligible} during inference. On the other hand, if we only shift a small proportion of the channels, \eg, $1/8$, we can limit the latency overhead to  \textbf{only 3\%}. Therefore, we use \emph{partial shift} strategy in our TSM implementation to significantly bring down the memory movement cost.

\myparagraph{Keeping Spatial Feature Learning Capacity.}

We need to balance the model capacity for spatial feature learning and temporal feature learning.
A straight-forward way to apply TSM is to insert it before each convolutional layer or residual block, as illustrated in Figure~\ref{fig:shift_inplace}. We call such implementation \emph{in-place shift}. It harms the spatial feature learning capability of the backbone model, especially when we shift a large amount of channels, since the information stored in the shifted channels is lost for the current frame.

To address such issue, we propose a variant of the shift module. Instead of inserting it in-place, we put the TSM \emph{inside} the residual branch in a residual block.
We denote such version of shift as \emph{residual shift} as shown in~\ref{fig:shift_residual}. Residual shift can address the degraded spatial feature learning problem, as all the information in the original activation is still accessible after temporal shift through identity mapping.

To verify our assumption, we compared the performance of in-place shift and residual shift on Kinetics~\cite{kay2017kinetics} dataset. We studied the experiments under different shift proportion setting. The results are shown in~\ref{fig:tsm_discuss_b}. We can see that residual shift achieves better performance than in-place shift for all shift proportion. Even we shift all the channels to neighboring frames, due to the shortcut connection, residual shift still achieves better performance than the 2D baseline.
Another finding is that the performance is related to the proportion of shifted channels: if the proportion is too small, the ability of temporal reasoning may not be enough to handle complicated temporal relationships; if too large, the spatial feature learning ability may be hurt. For residual shift, we found that the performance reaches the peak when $1/4$ ($1/8$ for each direction) of the channels are shifted. Therefore, we use this setting for the rest of the paper.

\input{figureText/fig_shift_place}

%% file: figureText/fig_tsm_discuss.tex
\begin{figure}[t]
\centering
\begin{subfigure}[b]{0.23\textwidth}
    	\includegraphics[width=\textwidth]{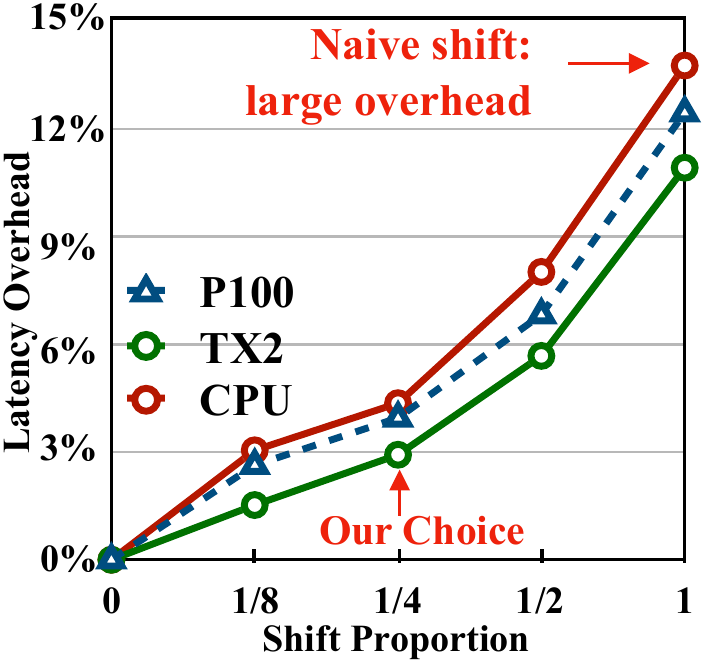}
	\caption{Overhead \vs proportion.}
	\label{fig:tsm_discuss_a}
\end{subfigure}
\begin{subfigure}[b]{0.23\textwidth}
	\includegraphics[width=\textwidth]{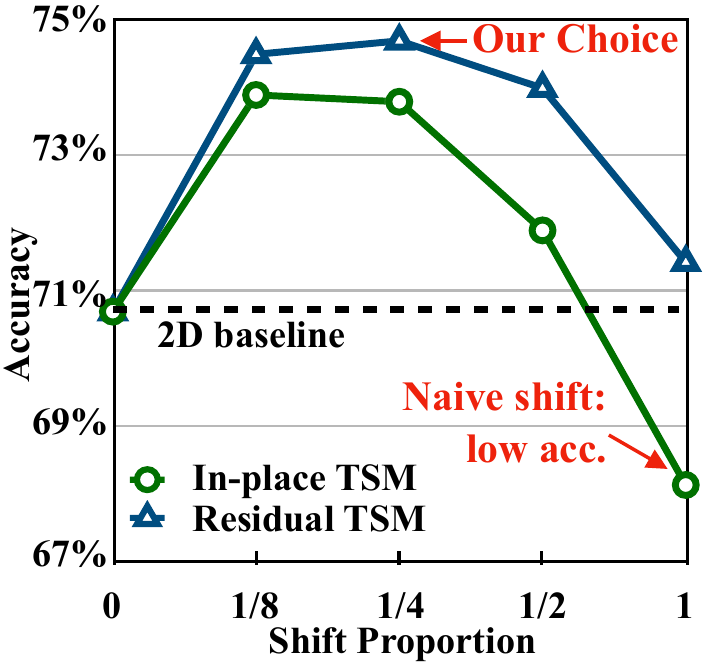}
	\caption{Residual \vs in-place.}
	\label{fig:tsm_discuss_b}
\end{subfigure}
\caption{\textbf{(a)} Latency overhead of TSM due to data movement. \textbf{(b)} Residual TSM achieve better performance than in-place shift. We choose 1/4 proportion residual shift  as our default setting. It achieves higher accuracy with a negligible overhead.}
\label{fig:tsm_discuss}
\vspace{-10pt}
\end{figure}

%% file: figureText/fig_shift_place.tex
\begin{figure}[t]
\centering
\begin{subfigure}[b]{0.22\textwidth}
	\includegraphics[width=\textwidth]{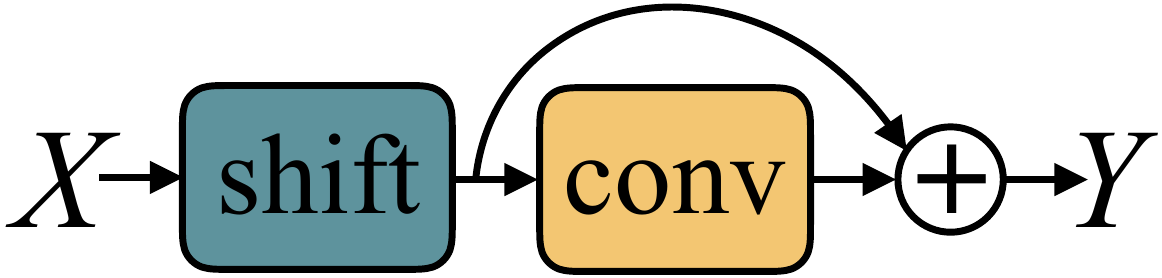}
	\caption{In-place TSM.}
	\label{fig:shift_inplace}
\end{subfigure}
~~
\begin{subfigure}[b]{0.22\textwidth}
	\includegraphics[width=\textwidth]{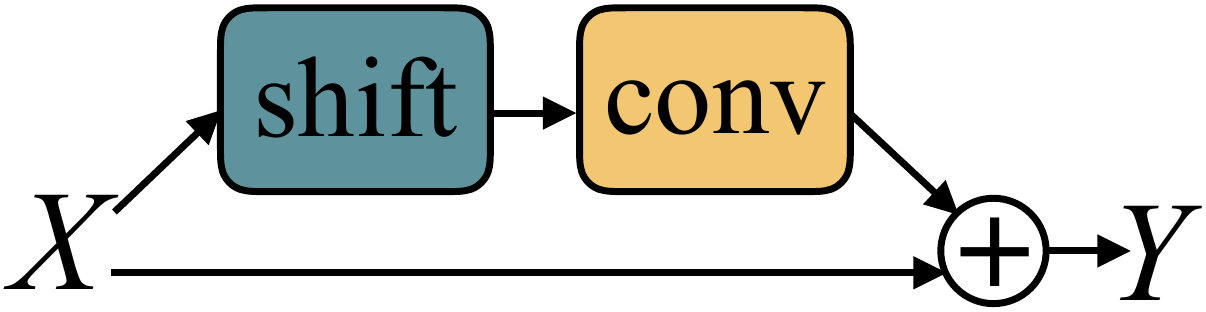}
	\caption{Residual TSM.}
	\label{fig:shift_residual}
\end{subfigure}
\caption{Residual shift is better than in-place shift. In-place shift happens before a convolution layer (or a residual block). Residual shift fuses temporal information inside a residual branch.}
\label{fig:shift_place}
\vspace{-20pt}
\end{figure}

%% file: text/4_tsm_video_network.tex
\section{TSM Video Network}

\subsection{Offline Models with Bi-directional TSM}
We insert bi-directional TSM to build offline video recognition models.
Given a video $V$, we first sample $T$ frames $F_i, F_1, ..., F_T$ from the video.
After frame sampling, 2D CNN baselines process each of the frames individually, and the output logits are averaged to give the final prediction. 
Our proposed \netFull model has exactly the same parameters and computation cost as 2D model. During the inference of convolution layers, the frames are still running independently just like the 2D CNNs. The difference is that  TSM is inserted for each residual block, which enables temporal information fusion at no computation.
For each inserted temporal shift module, the temporal receptive field will be enlarged by 2, as if running a convolution with the kernel size of 3 along the temporal dimension. Therefore, our \netFull model has a very large temporal receptive field to conduct highly complicated temporal modeling. In this paper, we used ResNet-50~\cite{he2016deep} as the backbone unless otherwise specified.

A unique advantage of TSM is that it can easily convert any off-the-shelf 2D CNN model into a pseudo-3D model that can handle both spatial and temporal information, without adding additional computation. Thus the deployment of our framework is hardware friendly: we only need to support the operations in 2D CNNs, which are already well-optimized at both framework level (CuDNN~\cite{chetlur2014cudnn}, MKL-DNN, TVM \cite{chen2018tvm}) and hardware level (CPU/GPU/TPU/FPGA).

\subsection{Online Models with Uni-directional TSM}

\input{figureText/fig_singlesided}

Video understanding from online video streams is important in real-life scenarios. Many real-time applications requires online video recognition with low latency, such as AR/VR and self-driving. In this section, we show that we can adapt TSM to achieve online video recognition while with multi-level temporal fusion.

As shown in Figure~\ref{fig:shift}, offline TSM shifts part of the channels bi-directionally, which requires features from future frames to replace the features in the current frame. If we only shift the feature from previous frames to current frames, we can achieve online recognition with uni-directional TSM.

The inference graph of uni-directional TSM for online video recognition is shown in Figure~\ref{fig:singlesided}. During inference, for each frame, we save the first 1/8 feature maps of each residual block and cache it in the memory. For the next frame, we replace the first 1/8 of the current feature maps with the cached feature maps. We use the combination of 7/8 current feature maps and 1/8 old feature maps to generate the next layer, and repeat. Using the uni-directional TSM for online video recognition shares several unique advantages: 

 \textbf{1. Low latency inference}. For each frame, we only need to replace and cache 1/8 of the features, without incurring any extra computations. Therefore, the latency of giving per-frame prediction is almost the same as the 2D CNN baseline. Existing methods like~\cite{zolfaghari2018eco} use multiple frames to give one prediction, which may leads to large latency.

\textbf{2. Low memory consumption}. Since we only cache a small portion of the features in the memory, the memory consumption is low. For ResNet-50, we only need 0.9MB memory cache to store the intermediate feature.

\textbf{3. Multi-level temporal fusion}. Most of the online method only enables late temporal fusion after feature extraction like~\cite{zhou2017temporal} or mid level temporal fusion~\cite{zolfaghari2018eco}, while our TSM enables all levels of temporal fusion. Through experiments (Table~\ref{tab:compare_something}) we find that multi-level temporal fusion is very important for complex temporal modeling. 

%% file: figureText/fig_singlesided.tex
\begin{figure}[t]
\centering
\vspace{-10pt}
\includegraphics[width=0.4\textwidth]{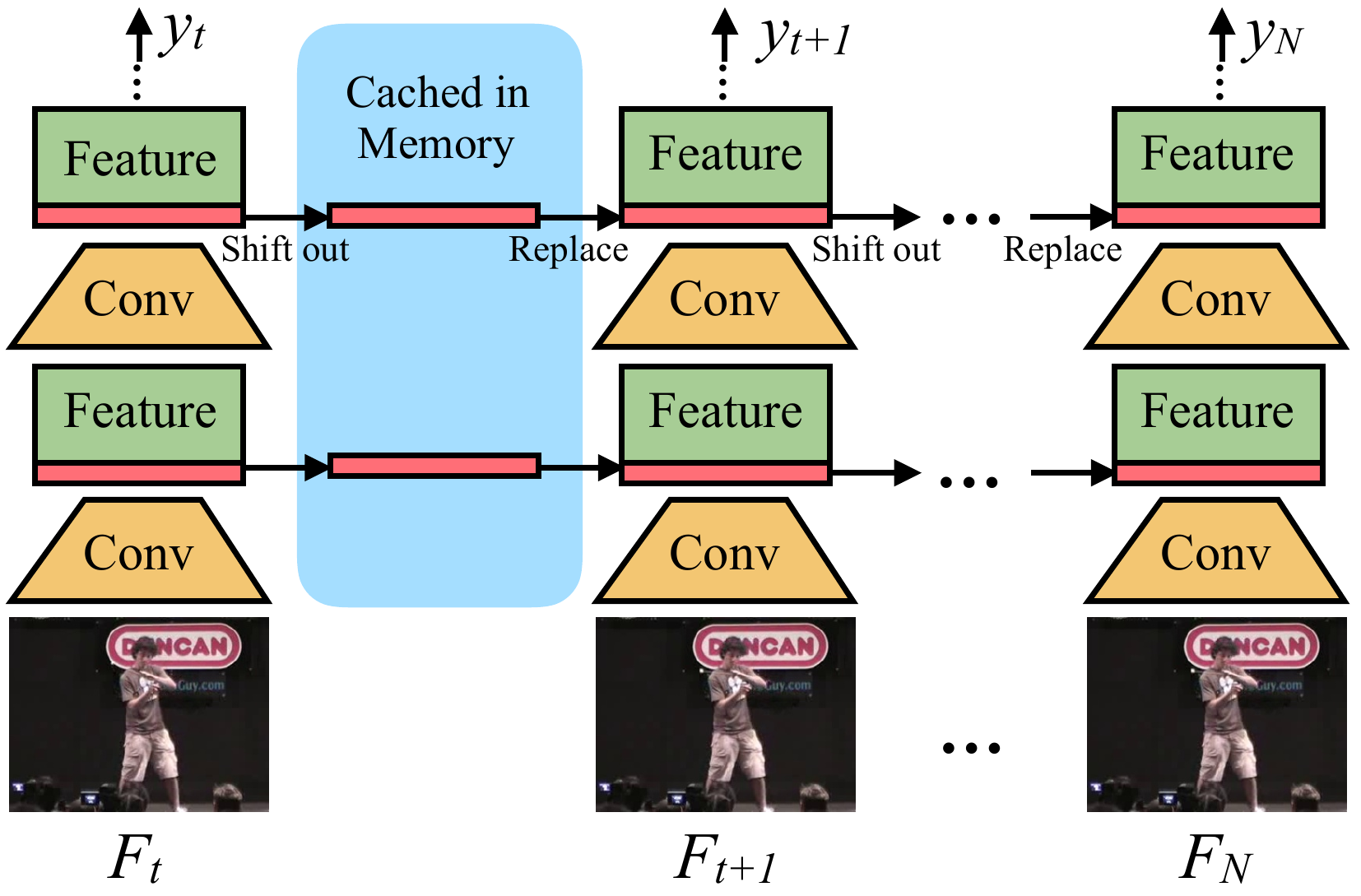}
\caption{Uni-directional TSM for online video recognition.}
\label{fig:singlesided}
\vspace{-15pt}
\end{figure}

%% file: text/5_experiments.tex
\section{Experiments}

We first show that TSM can significantly improve the performance of 2D CNN on video recognition while being computationally free and hardware efficient. It further demonstrated state-of-the-art performance on temporal-related datasets, arriving at a much better accuracy-computation pareto curve. TSM models achieve an order of magnitude speed up in measured GPU throughput compared to conventional I3D model from~\cite{wang2018videos}. Finally, we leverage uni-directional TSM to conduct low-latency and real-time online prediction on both video recognition and object detection.

\subsection{Setups}
\myparagraph{Training \& Testing.} 
We conducted experiments on video action recognition tasks.
The training parameters for the Kinetics dataset are: 100 training epochs, initial learning rate 0.01 (decays by 0.1 at epoch 40\&80), weight decay 1e-4, batch size 64, and dropout 0.5. For other datasets, we scale the training epochs by half.
For most of the datasets, the model is fine-tuned from ImageNet pre-trained weights; while HMDB-51~\cite{kuehne2011hmdb} and UCF-101~\cite{soomro2012ucf101} are too small and prone to over-fitting~\cite{wang2016temporal},
we followed the common practice~\cite{wang2016temporal, wang2017non} to 
fine-tune from Kinetics~\cite{kay2017kinetics} pre-trained weights and freeze the Batch Normalization~\cite{ioffe2015batch} layers.
For testing, when pursue high accuracy, we followed the common setting in~\cite{wang2017non, wang2018videos} to sample multiple clips per video (10 for Kinetics, 2 for others) and use the full resolution image with shorter side 256 for evaluation, so that we can give a direct comparison; when we consider the efficiency (\eg, as in Table~\ref{tab:compare_something}), we used just 1 clip per video and the center 224$\times$224 crop for evaluation. We keep the same protocol for the methods compared in the same table.

\myparagraph{Model.} 
To have an apple-to-apple comparison with the state-of-the-art method~\cite{wang2018videos}, we used the same backbone (ResNet-50) on the dataset ( Something-Something-V1~\cite{goyal2017something}).This dataset focuses on temporal modeling.  The difference is that \cite{wang2018videos} used 3D ResNet-50, while we used 2D ResNet-50 as the backbone to demonstrate efficiency.

\myparagraph{Datasets.}

Kinetics dataset~\cite{kay2017kinetics} is a large-scale action recognition dataset with 400 classes. As pointed in~\cite{zhou2017temporal, xie2018rethinking}, datasets like Something-Something (V1\&V2)~\cite{goyal2017something}, Charades~\cite{sigurdsson2016hollywood}, and Jester~\cite{jester}  are more focused on modeling the temporal relationships
, while UCF101~\cite{soomro2012ucf101}, HMDB51~\cite{kuehne2011hmdb}, and Kinetics~\cite{kay2017kinetics} are less sensitive to temporal relationships. Since TSM focuses on temporal modeling, we mainly focus on datasets with stronger temporal relationships like Something-Something. Nevertheless, we also observed strong results on the other datasets and reported it.

\subsection{Improving 2D CNN Baselines}

\input{figureText/tab_compare_2d_new_8f.tex}

\input{figureText/tab_compare_something}

We can seamlessly inject TSM into a normal 2D CNN  and improve its performance on video recognition. In this section, we demonstrate a 2D CNN baseline can significantly benefit from TSM with double-digits accuracy improvement.  
We chose TSN~\cite{wang2016temporal} as the 2D CNN baseline. We used the same training and testing protocol for TSN and our \netFull. The only difference is with or without TSM.  

\paragraph{Comparing Different Datasets. } We compare the results on several action recognition datasets in Table~\ref{tab:compared_2d}. The chart is split into two parts. The upper part contains datasets Kinetics~\cite{kay2017kinetics}, UCF101~\cite{soomro2012ucf101}, HMDB51~\cite{kuehne2011hmdb}, where temporal relationships are less important, while our \netFull still consistently outperforms the 2D TSN baseline at no extra computation. For the lower part, we present the results on Something-Something V1 and V2~\cite{goyal2017something} and Jester~\cite{jester}, which depend heavily on temporal relationships. 2D CNN baseline cannot achieve a good accuracy, but once equipped with TSM, the performance improved by double digits.

\input{figureText/tab_compare_backbone.tex}

\paragraph{Scaling over Backbones.} TSM scales well to backbones of different sizes. We show the Kinetics top-1 accuracy with MobileNet-V2~\cite{sandler2018mobilenetv2}, ResNet-50~\cite{he2016deep}, ResNext-101~\cite{xie2017aggregated} and ResNet-50 + Non-local module~\cite{wang2017non} backbones in Table~\ref{tab:compared_backbone}. TSM consistently improves the accuracy over different backbones, even for NL R-50, which already has temporal modeling ability.

\subsection{Comparison with State-of-the-Arts}

\netFull not only significantly improves the 2D baseline but also outperforms state-of-the-art methods, which heavily rely on 3D convolutions. We compared the performance of our \netFull model with state-of-the-art methods on both Something-Something V1\&V2 because these two datasets focus on temporal modeling.

\myparagraph{Something-Something-V1.} Something-Something-V1 is a challenging dataset, as activity cannot be inferred merely from individual frames (\eg, pushing something from \emph{right to left}).
We compared TSM with current state-of-the-art methods in Table~\ref{tab:compare_something}. 
We only applied center crop during testing to ensure the efficiency unless otherwise specified. TSM achieves \emph{the first place} on the leaderboard upon publication.

We first show the results of the 2D based methods TSN~\cite{wang2016temporal} and TRN~\cite{zhou2017temporal}. TSN with different backbones fails to achieve decent performance ($<$20\%  Top-1) due to the lack of temporal modeling. For TRN, although \textbf{late temporal fusion} is added after feature extraction, the performance is still significantly lower than state-of-the-art methods, showing the importance of temporal fusion across all levels. 

The second section shows the state-of-the-art efficient video understanding framework ECO~\cite{zolfaghari2018eco}. 
ECO uses an early 2D + late 3D architecture which enables \textbf{medium-level temporal fusion}.
Compared to ECO, our method achieves better performance at a smaller FLOPs. For example, when using 8 frames as input, our \netFull achieves 45.6\% top-1 accuracy with 33G FLOPs, which is 4.2\% higher accuracy than ECO with 1.9$\times$ less computation. The ensemble versions of ECO (ECO\textsubscript{\emph{En}}\emph{Lite} and ECO\textsubscript{\emph{En}}\emph{Lite}\textsubscript{RGB+Flow}, using an ensemble of  \{16, 20, 24, 32\} frames as input) did achieve competitive results, but the computation and parameters are too large for deployment. While our model is much more efficient: we only used \{8, 16\} frames model for ensemble (\netHead\textsubscript{\emph{En}}), and the model achieves better performance using 2.7$\times$ less computation and 3.1$\times$ fewer  parameters.

The third section contains previous state-of-the-art methods: Non-local I3D + GCN~\cite{wang2018videos}, that enables \textbf{all-level temporal fusion}. The GCN needs a Region Proposal Network~\cite{ren2015faster} trained on MSCOCO object detection dataset~\cite{lin2014microsoft} to generate the bounding boxes, which is unfair to compare since external data (MSCOCO) and extra training cost is introduced. Thus we compared \netFull to its CNN part: Non-local I3D. Our \netFull (8f) achieves 1.2\% better accuracy with $10\times$ fewer FLOPs on the validation set compared to the Non-local I3D network. Note that techniques like Non-local module~\cite{wang2017non} are orthogonal to our work, which could also be added to our framework to boost the performance further.

\myparagraph{Generalize to Other Modalities.} We also show that our proposed method can generalize to other modalities like optical flow. To extract the optical flow information between frames, we followed~\cite{wang2016temporal} to use the TVL1 optical flow algorithm~\cite{zach2007duality} implemented in OpenCV with CUDA. We conducted two-stream experiments on both Something-Something V1 and V2 datasets, and it consistently improves over the RGB performance: introducing optical flow branch brings 5.4\% and 2.6\% top-1 improvement on V1 and V2.

\input{figureText/tab_somethingv2}

\input{figureText/fig_acc_vs_flops.tex}

\myparagraph{Something-Something-V2.} We also show the result on Something-Something-V2 dataset, which is a newer release to its previous version. The results compared to other state-of-the-art methods are shown in Table~\ref{tab:somethingv2}. 
On Something-Something-V2 dataset, we achieved state-of-the-art performance while only using RGB input.

\myparagraph{Cost \vs Accuracy.}

Our \netFull model achieves very competitive performance while enjoying high efficiency and low computation cost for fast inference. We show the FLOPs for each model in Table~\ref{tab:compare_something}. 
Although GCN itself is light, the method used a ResNet-50 based Region Proposal Network~\cite{ren2015faster} to extract bounding boxes, whose cost is also considered in the chart.
Note that the computation cost of optical flow extraction is usually larger than the video recognition model itself. Therefore, we do not report the FLOPs of two-stream based methods.

We show the accuracy, FLOPs, and number of parameters trade-off in Figure~\ref{fig:acc_vs_flops}. The accuracy is tested on the validation set of Something-Something-V1 dataset, and the number of parameters is indicated by the area of the circles. We can see that our \netFull based methods have a better Pareto curve than both previous state-of-the-art efficient models (ECO based models) and high-performance models (non-local I3D based models). 
\netFull models are both efficient and accurate. It can achieve state-of-the-art accuracy at high efficiency: it achieves better performance while consuming $3\times$ less computation than the ECO family
. Considering that ECO is already an efficiency-oriented design, our method enjoys highly competitive hardware efficiency.

\subsection{Latency and Throughput Speedup}

\input{figureText/tab_runtime.tex}

The measured inference latency and throughput are important for the large-scale video understanding.
TSM has low latency and high throughput. We performed measurement on a single NVIDIA Tesla P100 GPU. We used batch size of 1 for latency measurement; batch size of16 for throughput measurement. 
We made two comparisons: 

(1) Compared with the I3D model from~\cite{wang2018videos}, our method is faster by an order of magnitude
at 1.8\% higher accuracy (Table~\ref{tab:runtime}). We also compared our method to the state-of-the-art efficient model ECO~\cite{zolfaghari2018eco}: Our \netFull model has $1.75\times$ lower latency (17.4ms \vs 30.6ms), $1.7\times$ higher throughput, and achieves 2\% better accuracy.  ECO has a two-branch (2D+3D) architecture, while TSM only needs the in-expensive 2D backbone.

(2) We then compared TSM to efficient 3D model designs.
One way is to only inflate the first $1\times1$ convolution in each of the block as in~\cite{wang2017non}, denoted as "I3D from~\cite{wang2017non}" in the table. 
Although the FLOPs are similiar due to pooling, it suffers from 1.5$\times$ higher latency and only 55\% the throughput compared with TSM, with worse accuracy. 
We speculate the reason is that TSM model only uses 2D convolution which is highly optimized for hardware.
To excliude the factors of backbone design, we replace every TSM primitive with $3\times1\times1$ convolution and denote this model as I3D\textsubscript{replace}. It is still much slower than TSM and performs worse.

\subsection{Online Recognition with TSM}

\myparagraph{Online \vs Offline}
Online TSM models shift the feature maps uni-directionally so that it can give predictions in real time. 
We compare the performance of offline and online TSM models to show that online TSM can still achieve comparable performance. 
Follow~\cite{zolfaghari2018eco}, we use the prediction averaged from all the frames to compare with offline models, \ie, we compare the performance after observing the whole videos. 
The performance is provided in Table~\ref{tab:offline_online}.
We can see that for less temporal related datasets like Kinetics, UCF101 and HMDB51, the online models achieve comparable and sometimes even better performance compared to the offline models. While for more temporal related datasets Something-Something, online model performs worse than offline model by 1.0\%. Nevertheless, the performance of online model is still significantly better than the 2D baseline.

\input{figureText/tab_online_vs_offline}

\input{figureText/fig_early_recognition}

We also compare the per-frame prediction latency of pure 2D backbone (TSN) and our online TSM model. We compile both models with TVM~\cite{chen2018tvm} on GPU.
Our online TSM model only adds to less than 0.1ms latency overhead per frame while bringing up to 25\% accuracy improvement. It demonstrates online TSM is hardware-efficient for latency-critical real-time applications.

\myparagraph{Early Recognition}

Early recognition aims to classify the video while only observing a small portion of the frames. It gives fast response to the input video stream. Here we compare the early video recognition performance on UCF101 dataset (Figure~\ref{fig:early_recognition}). Compared to ECO, TSM gives much higher accuracy, especially when only observing a small portion of the frames. For example, when only observing the first 10\% of video frames, TSM model can achieve 90\% accuracy, which is 6.6\% higher than the best ECO model.

\myparagraph{Online Object Detection}

Real-time online video object detection is an important application in self-driving vehicles, robotics, \etc. By injecting our online TSM into the backbone, we can easily take the temporal cues into consideration at negligible overhead, so that the model can handle poor object appearance like motion blur, occlusion, defocus, \etc.
We conducted experiments on R-FCN~\cite{dai16rfcn} detector with ResNet-101 backbone on ImageNet-VID~\cite{russakovsky2015imagenet} dataset. We inserted the uni-directional TSM to the backbone, while keeping other settings the same. The results are shown in Table~\ref{tab:detection}. 
Compared to 2D baseline R-FCN~\cite{dai16rfcn}, our online TSM model significantly improves the performance, especially on the fast moving objects, where TSM increases mAP by $4.6\%$. 
We also compare to a strong baseline FGFA~\cite{zhu2017flow} that uses optical flow to aggregate the temporal information from 21 frames (past 10 frames and future 10 frames) for offline video detection. Compared to FGFA, TSM can achieve similar or higher performance while enabling online recognition at much smaller latency. We visualize some video clips in the supplementary material to show that online TSM can leverage the temporal consistency to correct mis-predictions. 

\input{figureText/tab_detection.tex}

\myparagraph{Edge Deployment}
\input{figureText/tab_edge_deploy.tex}

TSM is mobile device friendly. 
We build an online TSM model with MobileNet-V2 backbone, which achieves 69.5\% accuracy on Kinetics. The latency and energy on NVIDIA Jetson Nano \& TX2, Raspberry Pi 4B, Samsung Galaxy Note8, Google Pixel-1 is shown in Table~\ref{tab:edge_deploy}. The models are compiled using TVM~\cite{chen2018tvm}. Power is measured with a power meter, subtracting the static power. TSM achieves low latency and low power on edge devices.

%% file: figureText/tab_compare_2d_new_8f.tex
\renewcommand \arraystretch{0.9}
\begin{table}[t]
\caption{Our method consistently outperforms 2D counterparts on multiple datasets at zero extra computation (protocol: ResNet-50 8f input, 10 clips for Kinetics, 2 for others, full-resolution).}
\vspace{-6pt}
\label{tab:compared_2d}
\small
\begin{center}
\begin{tabular}{cccccc}
\toprule
& \textbf{Dataset} &  \textbf{Model} &  \textbf{Acc1} &  \textbf{Acc5} &  \textbf{$\Delta$ Acc1}  \\ 
\toprule
\multirow{6}{*}{\rotatebox{90}{Less Temporal\quad}} & \multirow{2}{*}{Kinetics} &  TSN & 70.6 & 89.2 & \multirow{2}{*}{+3.5}  \\ 
 &  & Ours & \textbf{74.1} & \textbf{91.2} \\ \cmidrule(l){2-6}
& \multirow{2}{*}{UCF101}  & TSN & 91.7 &  99.2& \multirow{2}{*}{+4.2} \\ 
 &  & Ours & \textbf{95.9} & \textbf{99.7} \\ \cmidrule(l){2-6}
& \multirow{2}{*}{HMDB51}  & TSN & 64.7 & 89.9 & \multirow{2}{*}{+8.8} \\ 
 &  & Ours & \textbf{73.5} & \textbf{94.3} & \\ 
\midrule \midrule
\multirow{6}{*}{\rotatebox{90}{More Temporal\quad}}  & \multirow{2}{*}{\shortstack{Something\\V1}}  & TSN & 20.5  & 47.5 & \multirow{2}{*}{+28.0} \\ 
 &  & Ours & \textbf{47.3} & \textbf{76.2} \\ \cmidrule(l){2-6}
& \multirow{2}{*}{\shortstack{Something\\V2}}  &  TSN & 30.4 & 61.0 & \multirow{2}{*}{+31.3}\\ 
 &  & Ours & \textbf{61.7} & \textbf{87.4} \\ \cmidrule(l){2-6}
& \multirow{2}{*}{Jester}  & TSN & 83.9 & 99.6 & \multirow{2}{*}{+11.7} \\ 
 &  & Ours & \textbf{97.0} & \textbf{99.9} \\ 
\bottomrule
\end{tabular}
\end{center}
\vspace{-15pt}
\end{table}

%% file: figureText/tab_compare_something.tex
\renewcommand \arraystretch{1.}
\begin{table*}[t]
\vspace{-5pt}
\setlength{\tabcolsep}{5pt}
\caption{Comparing \netFull against other methods on Something-Something dataset (center crop, 1 clip/video unless otherwise specified).  }
\vspace{-15pt}
\label{tab:compare_something}
\small
\begin{center}
\begin{tabular}{cccccccc}
\toprule
\textbf{Model}
& \textbf{Backbone} & \textbf{\#Frame} & \textbf{FLOPs/Video} & \textbf{\#Param.} & \textbf{Val Top-1} & 
\textbf{Val Top-5} & \textbf{Test Top-1}  \\ 
\toprule
TSN~\cite{zhou2017temporal} & BNInception & 8  &16G & 10.7M & 19.5 & - & - \\ 
TSN (our impl.) & ResNet-50 & 8  & 33G & 24.3M & 19.7 & 46.6 & -   \\ 
TRN-Multiscale~\cite{zhou2017temporal} & BNInception & 8 &16G & 18.3M & 34.4 & - & 33.6 \\ 
TRN-Multiscale (our impl.) & ResNet-50 & 8 &33G & 31.8M & 38.9 & 68.1  & - \\ 
Two-stream TRN\textsubscript{RGB+Flow}~\cite{zhou2017temporal} & BNInception & 8+8 & - & 36.6M & 42.0 & - & 40.7 \\ \midrule
ECO~\cite{zolfaghari2018eco} & BNIncep+3D Res18  & 8  & 32G & 47.5M & 39.6 & - & - \\
ECO~\cite{zolfaghari2018eco} & BNIncep+3D Res18 & 16 & 64G & 47.5M & 41.4 & - & - \\ 
ECO\textsubscript{\emph{En}}\emph{Lite}~\cite{zolfaghari2018eco} & BNIncep+3D Res18 & 92 & 267G & 150M & 46.4 & - & 42.3   \\
ECO\textsubscript{\emph{En}}\emph{Lite}\textsubscript{RGB+Flow}~\cite{zolfaghari2018eco} & BNIncep+3D Res18 & 92+92 & - & 300M & 49.5 & - & 43.9 \\ \midrule
I3D from~\cite{wang2018videos} & 3D ResNet-50 & 32$\times$2clip & 153G\tablefootnote{\label{note1} We reported the performance of NL I3D described in~\cite{wang2018videos}, which is a variant of the original NL I3D~\cite{wang2017non}. It uses fewer temporal dimension pooling to achieve good performance, but also incur larger computation.}$\times$2  & 28.0M & 41.6 & 72.2 & - \\ 
Non-local I3D from~\cite{wang2018videos} & 3D ResNet-50 & 32$\times$2clip & 168G\footnoteref{note1}$\times$2 & 35.3M & 44.4 & 76.0 & - \\  
Non-local I3D + GCN~\cite{wang2018videos} & 3D ResNet-50+GCN & 32$\times$2clip & 303G\tablefootnote{\label{note2}Includes parameters and FLOPs of the Region Proposal Network.}$\times$2 & 62.2M\footnoteref{note2} & 46.1 & 76.8  & 45.0\\  \midrule\midrule
\netFull & ResNet-50 & 8 & 33G & 24.3M & 45.6 & 74.2 & - \\
\netFull & ResNet-50 & 16 & 65G & 24.3M & 47.2 & 77.1 & 46.0 \\ \midrule
\netHead\textsubscript{\emph{En}} & ResNet-50 & 24 & 98G & 48.6M & 49.7 & 78.5 & - \\
\netHead\textsubscript{RGB+Flow} & ResNet-50 & 16+16 & - & 48.6M & \textbf{52.6} & \textbf{81.9}  & \textbf{50.7}\\
\bottomrule
\end{tabular}
\end{center}
\vspace{-15pt}
\end{table*}

%% file: figureText/tab_compare_backbone.tex
\renewcommand \arraystretch{0.9}
\begin{table}[t]
\caption{TSM can consistently improve the performance over different backbones on Kinetics dataset.}
\vspace{-6pt}
\label{tab:compared_backbone}
\small
\begin{center}
\begin{tabular}{ccccc}
\toprule
& \textbf{Mb-V2} &  \textbf{R-50} &  \textbf{RX-101} &  \textbf{NL R-50}\\ 
\midrule
TSN & 66.5 & 70.7 & 72.4 & 74.6 \\
TSM & 69.5 & 74.1 & 76.3 & 75.7 \\
\midrule
$\Delta$Acc. & +3.0 & +3.4 & +3.9 & +1.1  \\
\bottomrule 
\end{tabular}
\end{center}
\vspace{-20pt}
\end{table}

%% file: figureText/tab_somethingv2.tex
\renewcommand \arraystretch{0.95}
\begin{table}[t]
\setlength{\tabcolsep}{7pt}
\vspace{-20pt}
\caption{Results on Something-Something-V2. Our \netFull achieves state-of-the-art performance.}
\vspace{-10pt}
\label{tab:somethingv2}
\small
\begin{center}
\begin{tabular}{ccccc}
\toprule
\multirow{2}{*}{\textbf{Method}} & \multicolumn{2}{c}{\textbf{Val}}   &  \multicolumn{2}{c}{\textbf{Test}} \\
\cmidrule(lr){2-3}\cmidrule(lr){4-5}
& \textbf{Top-1} & \textbf{Top-5}   &  \textbf{Top-1} & \textbf{Top-5} \\ \midrule
TSN (our impl.) & 30.0 & 60.5 & - & - \\
MultiScale TRN~\cite{zhou2017temporal} & 48.8 & 77.6 & 50.9 & 79.3 \\
2-Stream TRN~\cite{zhou2017temporal} & 55.5 & 83.1 & 56.2 & 83.2 \\
\midrule
\netHead\textsubscript{8F} & 59.1 & 85.6 & - & - \\
\netHead\textsubscript{16F} & 63.4 & 88.5  & 64.3 & 89.6 \\
\netHead\textsubscript{RGB+Flow} & \textbf{66.0} & \textbf{90.5} & \textbf{66.6} & \textbf{91.3}\\ 
\bottomrule
\end{tabular}
\end{center}
\vspace{-10pt}
\end{table}

%% file: figureText/fig_acc_vs_flops.tex
\begin{figure}[t]
\centering
\vspace{-5pt}
\includegraphics[width=0.4\textwidth]{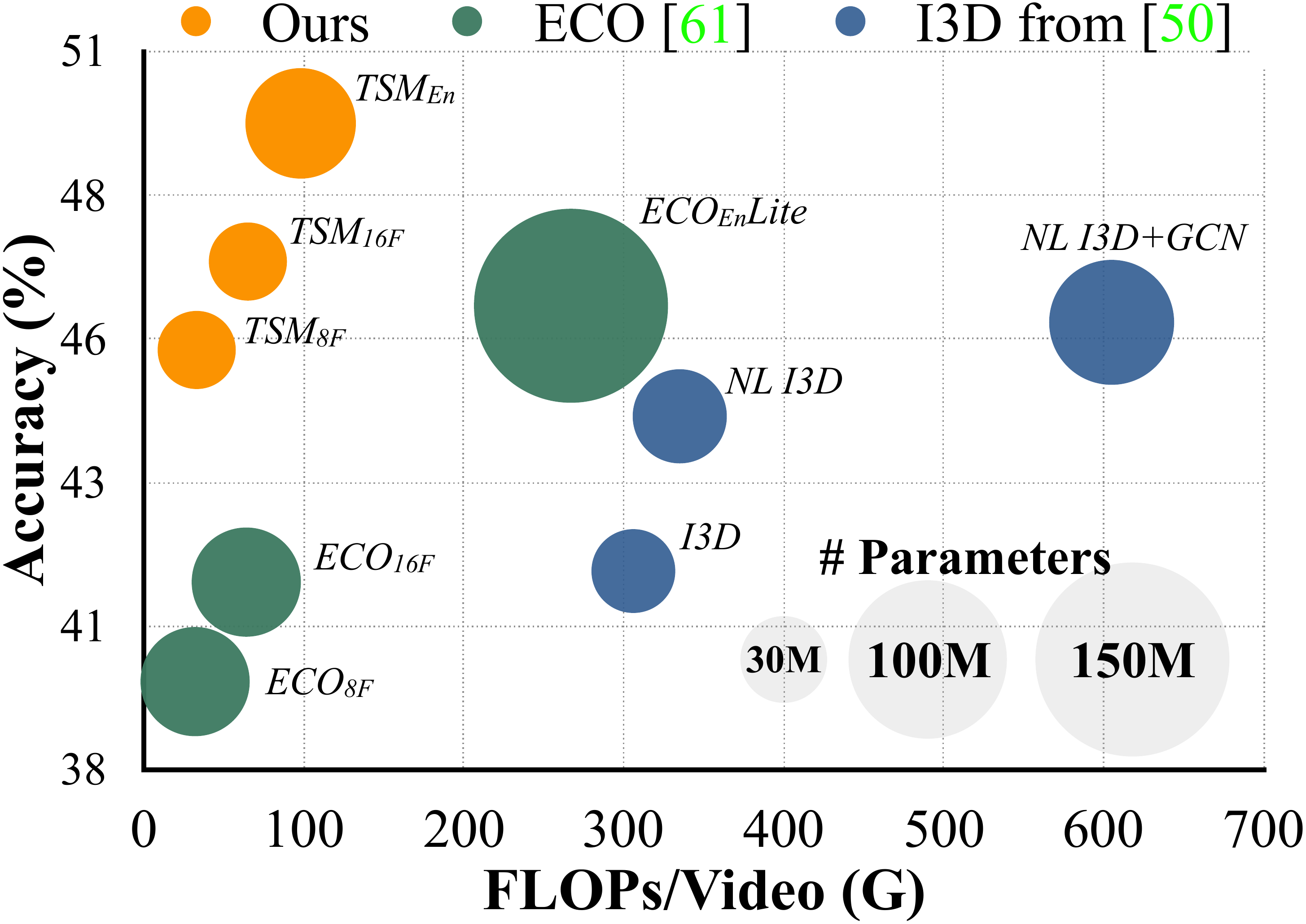}
\caption{
TSM enjoys better accuracy-cost trade-off than I3D family and ECO family on Something-Something-V1~\cite{goyal2017something} dataset.
(GCN includes the cost of ResNet-50 RPN to generate region proposals.)}
\vspace{-10pt}
\label{fig:acc_vs_flops}
\end{figure}

%% file: figureText/tab_runtime.tex
\renewcommand \arraystretch{1.2}
\begin{table}[t]
\vspace{-20pt}
\setlength{\tabcolsep}{2.5pt}
\caption{TSM enjoys low GPU inference latency and high throughput.
V/s means videos per second, higher the better (Measured on NVIDIA Tesla P100 GPU). }
\vspace{-10pt}
\label{tab:runtime}
\begin{center}
\footnotesize{
    \begin{tabular}{ccccccc}
    \toprule
    \multirow{2}{*}{\textbf{Model}} &
     \multicolumn{4}{c}{\textbf{Efficiency Statistics}}  & \multicolumn{2}{c}{\textbf{Accuracy}} \\ 
     \cmidrule(lr){2-5}\cmidrule(lr){6-7}
     &  \textbf{FLOPs} & \textbf{Param.} &  \textbf{Latency} & \textbf{Thrput.} & \textbf{Sth.} & \textbf{Kinetics} \\ \toprule
    I3D from~\cite{wang2018videos} & 306G & 35.3M & 165.3ms & 6.1V/s & 41.6\% & -\\  
    ECO\textsubscript{16F}~\cite{zolfaghari2018eco}& 64G & 47.5M  & 30.6ms & 45.6V/s & 41.4\% & - \\ 
    \midrule
    I3D from~\cite{wang2017non} & \textbf{33G} & 29.3M & 25.8ms & 42.4V/s & - & 73.3\% \\
    I3D\textsubscript{replace} & 48G & 33.0M & 28.0ms & 37.9V/s & 44.9\% & -\\
    \midrule
     \netHead\textsubscript{8F}  & \textbf{33G} & \textbf{24.3M} & \textbf{17.4ms} & \textbf{77.4V/s} & 45.6\% & 74.1\%  \\ 
    \netHead\textsubscript{16F}  & 65G & 24.3M & 29.0ms & 39.5V/s & \textbf{47.2\%} & \textbf{74.7\%} \\ 
    \bottomrule
    \end{tabular}
}
\end{center}
\vspace{-25pt}
\end{table}

%% file: figureText/tab_online_vs_offline.tex
\renewcommand \arraystretch{1.}
\begin{table}[t]
\setlength{\tabcolsep}{3pt}
\vspace{-20pt}
\caption{Comparing the accuracy of offline TSM and online TSM on different datasets. Online TSM brings negligible latency overhead.}
\vspace{-15pt}
\label{tab:offline_online}
\small
\begin{center}
\begin{tabular}{cccccc}
\toprule
\textbf{Model} & \textbf{Latency}
& \textbf{Kinetics} & \textbf{UCF101} & \textbf{HMDB51} & \textbf{Something}  \\ 
\toprule
TSN & 4.7ms & 70.6\%  & 91.7\% & 64.7\% & 20.5\% \\ \midrule
+Offline & - & 74.1\% &  95.9\% & 73.5\% & 47.3\% \\ 
+Online & 4.8ms & 74.3\% & 95.5\% & 73.6\%  & 46.3\% \\
\bottomrule
\end{tabular}
\end{center}
\vspace{-13pt}
\end{table}

%% file: figureText/fig_early_recognition.tex
\begin{figure}[t]
\centering
\includegraphics[width=0.38\textwidth]{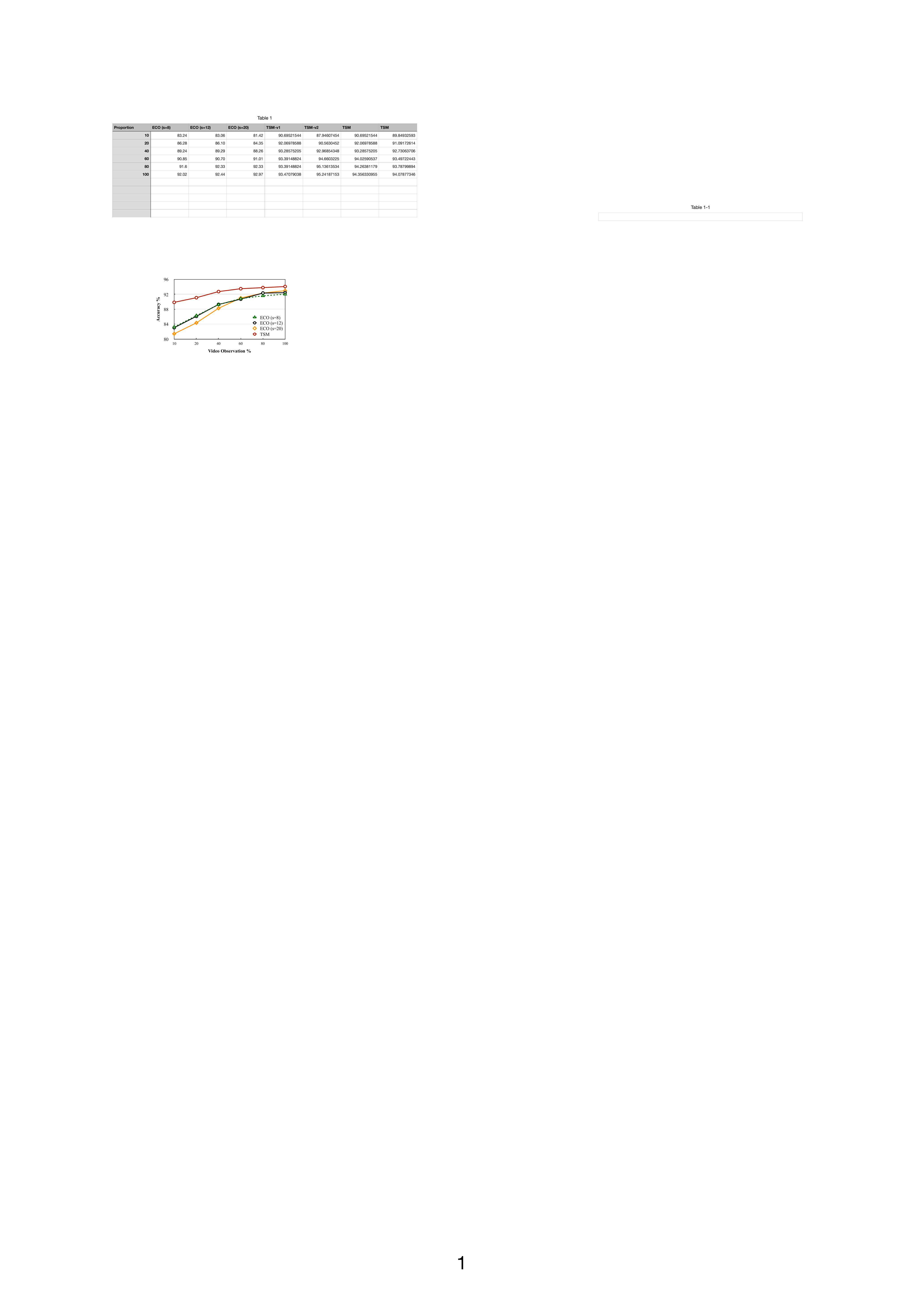}
\caption{Early recognition on UCF101.  TSM gives high prediction accuracy after only observing a small portion of the video.}
\label{fig:early_recognition}
\vspace{-20pt}
\end{figure}

%% file: figureText/tab_detection.tex
\renewcommand \arraystretch{0.9}
\begin{table}[t]
\setlength{\tabcolsep}{2.5pt}
\vspace{-20pt}
\caption{Video detection results on ImageNet-VID. }
\label{tab:detection}
\begin{center}
\footnotesize{
    \begin{tabular}{cccccccc}
    \toprule
    \multirow{2}{*}{\textbf{Model}} & \multirow{2}{*}{\textbf{Online}} & \multirow{2}{*}{\textbf{\shortstack{Need\\Flow}}} & \multirow{2}{*}{\textbf{Latency}} &
    \multicolumn{4}{c}{\textbf{mAP}} \\ 
    \cmidrule(lr){5-8}
    & & & & \textbf{Overall} & \textbf{Slow} & \textbf{Medium} & \textbf{Fast} \\
    \midrule
    R-FCN~\cite{dai16rfcn} & \checkmark & & 1$\times$ & 74.7 & 83.6 & 72.5 & 51.4\\
    FGFA~\cite{zhu2017flow} & & \checkmark & 2.5$\times$  & 75.9 & \textbf{84.0} & 74.4 & 55.6  \\
    \midrule
    Online TSM & \checkmark &  & 1$\times$ & \textbf{76.3}& 83.4 & \textbf{74.8} & \textbf{56.0}\\
    \bottomrule
    \end{tabular}
}
\end{center}
\vspace{-15pt}
\end{table}

%% file: figureText/tab_edge_deploy.tex
\renewcommand \arraystretch{1}
\begin{table}[t]
\setlength{\tabcolsep}{3pt}
\caption{TSM efficiently runs on edge devices with low latency.}
\vspace{-10pt}
\label{tab:edge_deploy}
\small
\begin{center}
\begin{tabular}{cccccccc}
\toprule
\multirow{2}{*}{\textbf{Devices}} & \multicolumn{2}{c}{Jetson Nano} & \multicolumn{2}{c}{Jetson TX2} & \multirow{2}{*}{Rasp.} &   \multirow{2}{*}{Note8} & \multirow{2}{*}{Pixel1}  \\ \cmidrule(lr){2-3} \cmidrule(lr){4-5}
& CPU & GPU & CPU & GPU& \\
\toprule
\textbf{Latency} (ms) & 47.8 & 13.4 & 36.4 & 8.5 & 69.6  &34.5 & 47.4\\
\textbf{Power} (watt) & 4.8  & 4.5 & 5.6 & 5.8 & 3.8 & - & - \\
\bottomrule
\end{tabular}
\end{center}
\vspace{-25pt}
\end{table}

%% file: text/6_conclusion.tex
\vspace{-7pt}
\section{Conclusion}
\vspace{-7pt}
We propose Temporal Shift Module for hardware-efficient video recognition. It can be inserted into 2D CNN backbone to enable joint spatial-temporal modeling at no additional cost. The module shifts part of the channels along temporal dimension to exchange information with neighboring frames. Our framework is both efficient and accurate, enabling low-latency video recognition on edge devices.  

%% file: text/7_acknowledge.tex
\paragraph{Acknowledgments}
We thank MIT Quest for Intelligence, MIT-IBM Watson AI Lab, MIT-SenseTime Alliance, Samsung, SONY, AWS, Google for supporting this research. We thank Oak Ridge National Lab for Summit supercomputer.

%% file: text/appendix.tex
\newpage
\appendix
\newpage

\section{Uni-directional TSM for Online Video Detection}

In this section, we show more details about the online video object detection with uni-directional TSM.

Object detection suffers from poor object appearance due to motion blur, occlusion, defocus, \etc. Video based object detection gives chances to correct such errors by aggregating and inferring temporal information. 

Existing methods on video object detection ~\cite{zhu2017flow} fuses information along temporal dimension after the feature is extracted by the backbone. 
Here we show that we can enable temporal fusion in online video object detection by injecting our uni-directional TSM into the backbone. We show that we can significantly improve the performance of video detection by simply modifying the backbone with online TSM, without changing the detection module design or using optical flow features.

We conducted experiments with R-FCN~\cite{dai16rfcn} detector on ImageNet-VID~\cite{russakovsky2015imagenet} dataset. Following the setting in~\cite{zhu2017flow}, we used ResNet-101~\cite{he2016deep} as the backbone for R-FCN detector. For TSM experiments, we inserted uni-directional TSM to the backbone, while keeping other settings the same. We used the official training code of~\cite{zhu2017flow} to conduct the experiments, and the results are shown in Table~\ref{tab:detection}. 
Compared to 2D baseline R-FCN~\cite{dai16rfcn}, our online TSM model significantly improves the performance, especially on the fast moving objects, where TSM increases mAP by $4.6\%$. 
FGFA~\cite{zhu2017flow} is a strong baseline that uses optical flow to aggregate the temporal information from 21 frames (past 10 frames and future 10 frames) for offline video detection. Compared to FGFA, TSM can achieve similar or higher performance while enabling online recognition (using information from only past frames) at much smaller latency per frame.
The latency overhead of TSM module itself is less than 1ms per frame, making it a practical tool for real deployment.

\begin{figure*}[t]
\centering
\includegraphics[width=0.95\textwidth]{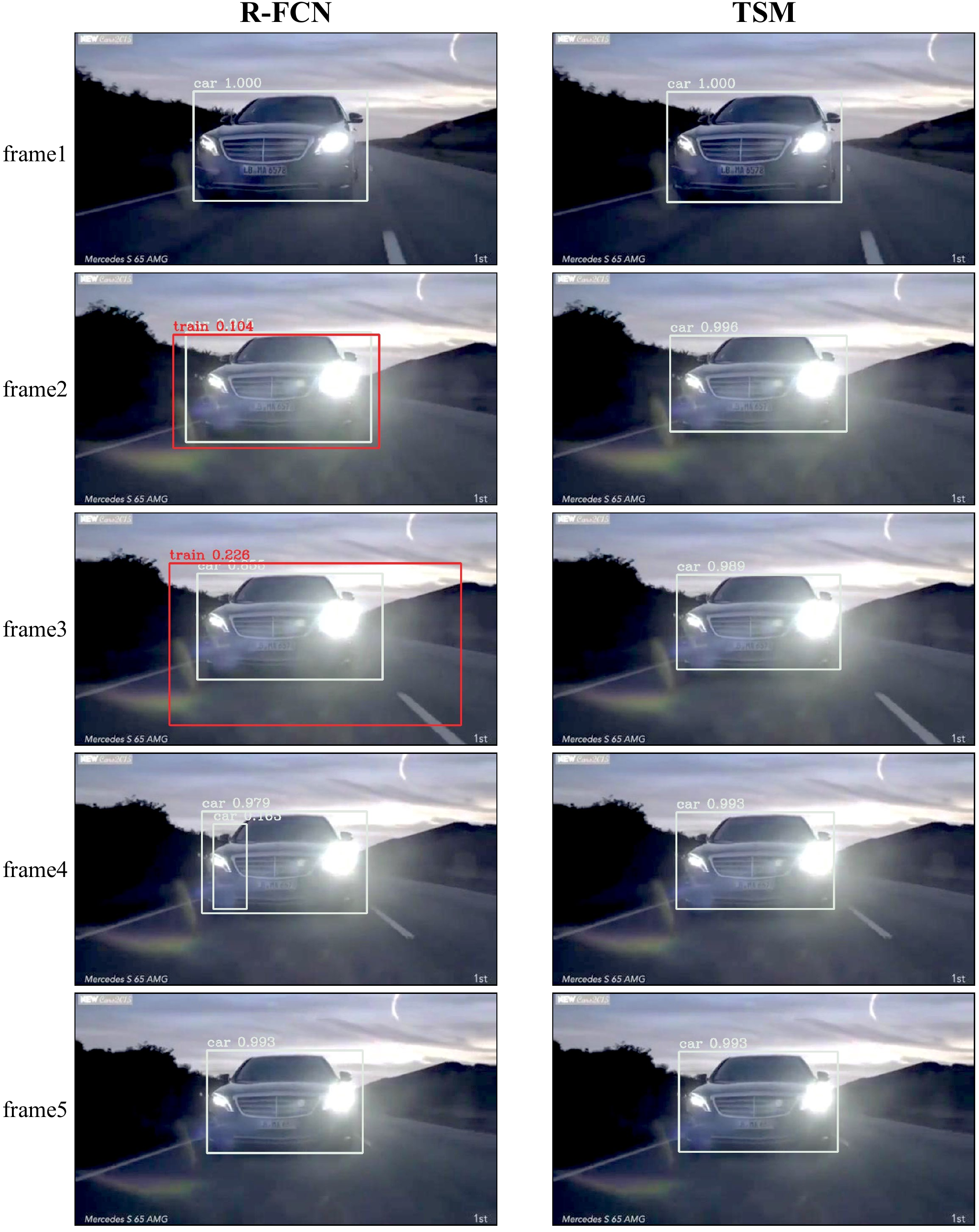}
\caption{Comparing the result of R-FCN baseline and TSM model. 2D baseline R-FCN generates false positive due to the glare of car headlight on frame 2/3/4, while TSM does not have such issue by considering the temporal information. }
\label{fig:det_car}
\end{figure*}

\begin{figure*}[t]
\centering
\includegraphics[width=0.95\textwidth]{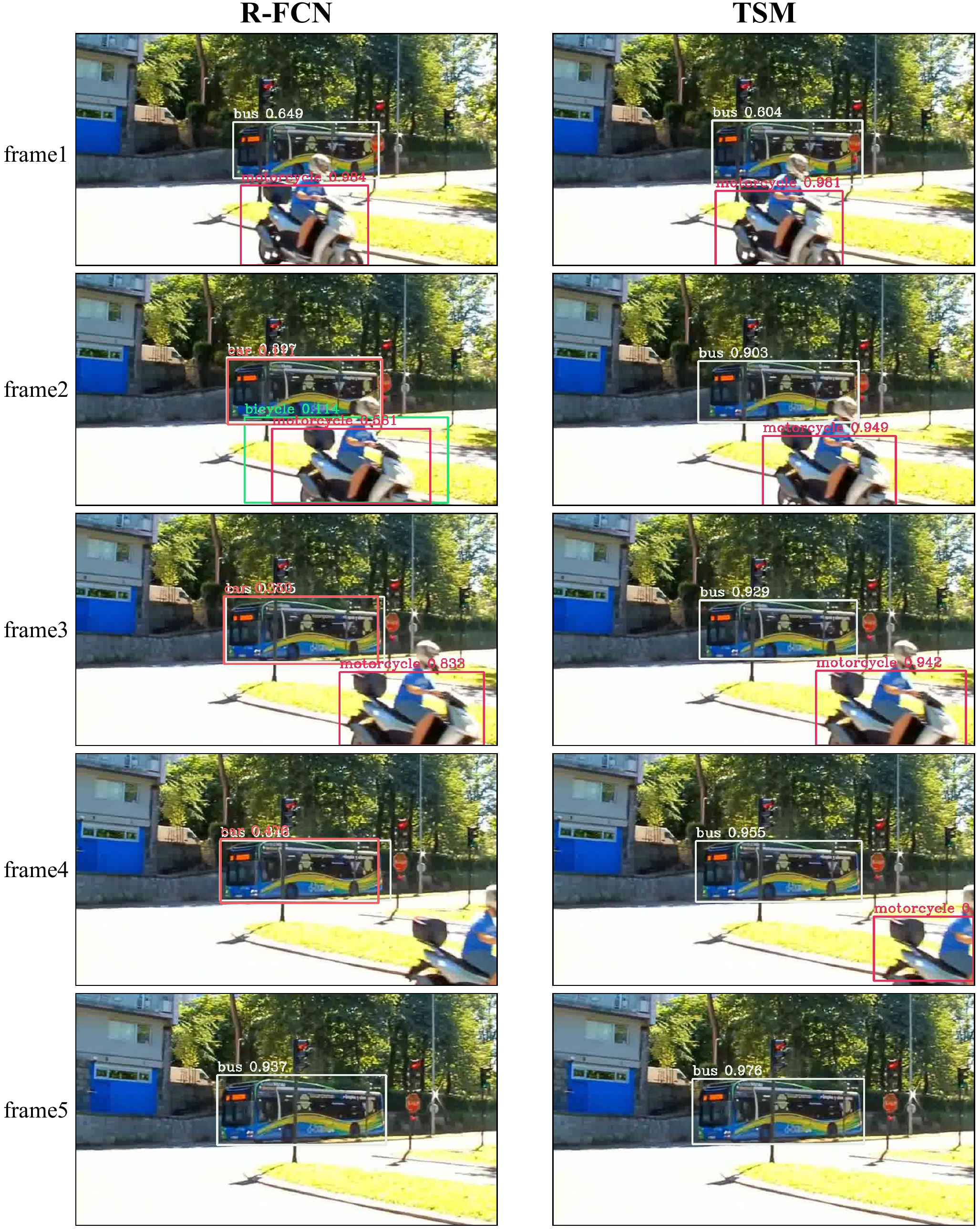}
\caption{Comparing the result of R-FCN baseline and TSM model. 2D baseline R-FCN generates false positive surrounding the bus due to occlusion by the traffic sign on frame 2/3/4. Also, it fails to detect motorcycle on frame 4 due to occlusion. TSM model addresses such issues with the help of temporal information. }
\label{fig:det_bus}
\end{figure*}

We visualize two video clips in Figure~\ref{fig:det_car} and~\ref{fig:det_bus}. In Figure~\ref{fig:det_car}, 2D baseline R-FCN generates false positive due to the glare of car headlight on frame 2/3/4, while TSM suppresses false positive. In Figure~\ref{fig:det_bus}, R-FCN generates false positive surrounding the bus due to occlusion by the traffic sign on frame 2/3/4. Also, it fails to detect motorcycle on frame 4 due to occlusion. TSM model addresses such issues with the help of temporal information.

\section{Video Demo}

We provide more video demos of our TSM model in the following project page: \url{https://hanlab.mit.edu/projects/tsm/}.

\newpage

\newpage

